# Biological Inspiration for Artificial Immune Systems

Jamie Twycross and Uwe Aickelin

School of Computer Science, University of Nottingham, UK
`jpt@cs.nott.ac.uk`

**Abstract.** Artificial immune systems (AISs) to date have generally been inspired by naive biological metaphors. This has limited the effectiveness of these systems. In this position paper two ways in which AISs could be made more biologically realistic are discussed. We propose that AISs should draw their inspiration from organisms which possess only innate immune systems, and that AISs should employ systemic models of the immune system to structure their overall design. An outline of plant and invertebrate immune systems is presented, and a number of contemporary systemic models are reviewed. The implications for interdisciplinary research that more biologically-realistic AISs could have is also discussed.

## 1 Introduction

The field of Artificial Immune Systems began in the early 1990s with a number of independent groups conducting research which used the biological immune system as inspiration for solutions to problems in non-biological domains. Since that time artificial immune system (AIS) research has produced a considerable body of knowledge and a number of general purpose algorithms. AISs based on these algorithms have been applied to many benchmark and a number of real-world problems. Currently however, the field is at an impasse [1, 2]. While there have been some success stories on realworld problems, there is still little to differentiate the performance of AISs with other state-of-the-art methods. We concur with Timmis [2] that this is due to a limited application to challenging problems, a lack of theoretical advances, and the use of naive biological metaphors. In this position paper we focus on biological metaphors and discuss the areas of biology that we believe should be important in inspiring future AISs. Our intention is to draw the attention of AIS researchers to these areas and to provide references to key papers which we have found useful in understanding the biology.

This paper argues that AISs can be made more biologically realistic in two ways. In the first place, we believe that AIS researchers should consider drawing inspiration from simpler biological systems than humans. A serious evaluation of the validity and usefulness of building AISs inspired by the adaptive immune system needs to take place. The vast majority of life survives and flourishes without an adaptive immune system. The innate immune system mechanisms employed by the majority of organisms provide robust maintenance of organism integrity and protection against pathogens. While complex, these purely innate immune systems are relatively simpler in organisational terms than immune systems which combine both innate and adaptive arms. Recent research has also shown that innate immune systems exhibit properties such as specificity, diversity and memory, previously only associated with adaptive immune systems. Innate immune systems can and do everything adaptive immune systems do, including adapt to rapidly evolving pathogens, albeit using different mechanisms [3]. It seems only sensible to start with simpler innate-based AISs before building adaptive immune mechanisms into AISs.

Secondly, AISs need to be based around more contemporary and sophisticated systemic models of the immune system than those currently employed. As shown by these contemporary models, the view of the immune system as a protective system driven by adaptive immune system mechanisms of self/nonself discrimination is at odds with current immunological thinking on how the immune system behaves as a complete system. While self/nonself discrimination is a characteristic observed in both innate and adaptive immune systems, it is not the purpose of the immune system. Yet, as a survey of past ICARIS proceedings [4] reveals, the majority of AISs built so far have been built for the purpose of discriminating self from nonself. This is not just arguing over semantics, but goes to the heart of the engineering philosophy used to build AISs.

Even if we must build AISs which incorporate adaptive immune system mechanisms, it makes little sense to build them based only the adaptive immune system. There is no organism in existence with only an adaptive immune system. Organisms which do possess an adaptive immune system also have innate immune systems. There seems to us to be a very good reason for this. While the adaptive immune system provides the organism with a diverse set of receptors which can recognise almost any molecule, it provides very little control over this recognition. The control of the adaptive immune system is firmly in the hands of the innate immune system [5]. Building AISs which model only adaptive immune system mechanisms is like building a car without a steering wheel - it will certainly go somewhere, but you have very little control as to where this is!

In the first part of this paper we discuss current understanding of the immune systems of plants and invertebrates with the idea that these organisms could provide simpler biological systems from which to draw inspiration for AISs. In the second part of this paper we discuss systemic models of the human immune system. In particular, in light of the first part of this paper and the importance of the innate immune system, we focus on systemic models which are concerned with how the innate and adaptive immune systems are integrated. The paper ends with a brief discussion of the implications for interdisciplinary research that more biologically-realistic AISs could have.

## 2  Non-human immune systems

The majority of AISs to date have been inspired by vertebrate adaptive immune system mechanisms. This focus of AIS research on the adaptive immune system is in some ways similar to Artificial Intelligence's early focus on the human mind and symbolic information processing. Only more recently has the focus of AI been expanded by the acknowledgement of intelligence in the wider sense of adaptive behaviour of organisms other than humans. We firmly believe that the field of AISs also needs to reassess its sources of biological inspiration and focus on the immune systems possessed by the majority of life on this planet. The adaptive immune system may be interesting and useful, but is in no way a prerequisite for a successful immune system, just as playing chess is interesting and useful but is in no way a hallmark of intelligent behaviour. By studying plant and invertebrate immune systems, differences and commonalities that exist between immune systems can also be uncovered. This could well help identify general principles of immune systems which could be of use to AIS researchers.

2.1 Plant immune systems

Plants do not have specialised defender cells and rely on innate immunity provided by each cell in the plant. Upon infection with a pathogen, plant cells are induced to produce a range of antimicrobial products which help neutralise pathogens. Pathogens which survive usually trigger a hypersensitive cell death response (HR), which causes host cells at the site of infection to die. Both HR and production of antimicrobial products need to be tightly controlled and plants have evolved intricate systems to do this. Inducible plant immunity is provided by two different but interacting systems. The first system is based around pattern recognition receptors (PRRs) on the surface of plant cells. These PRRs are activated by molecules produced by pathogens called pathogen- or microbial-associate molecular proteins (PAMPs or MAMPs). The second system, which acts intracellularly, is based around a set of polymorphic proteins called nucleotide-binding site plus leucine-rich repeat (NB-LRR) proteins. These NB-LRR proteins are coded for in the genome of the plant by specific disease resistance (R) genes [6].

Inducible immunity in plants is currently viewed as a four-phase process. Phase 1 is initiated by the recognition of PAMPs or MAMPs by PRRs and induces a set of responses known as PAMP-triggered immunity (PTI). In Phase 2, pathogens which succeed in overcoming the initial PTI response produce effector molecules (also known as virulence factors) which enhance the spread of the pathogen and can also suppress PTI responses. Phase 2 results in effector-triggered susceptibility of the host to the pathogen. In Phase 3, effectors produced by the pathogen are recognised by NB-LRR proteins encoded by R genes and initiate effector-triggered immunity (ETI). The specific effector which is recognised is termed an avirulence (Avr) protein. ETI responses are amplified versions of PTI responses and usually result in the death of the infected host cell. In Phase 4 both host and pathogen undergo a process of selection in which pathogen variants which do not produce the triggering Avr protein but instead produce other effectors are selected for. At the same time host R genes which produce NB-LRR proteins which recognise the new effectors are selected for, once again resulting in ETI [6, 7].

While direct recognition of Avr proteins by NB-LRR proteins has been observed, indirect recognition of Avr proteins also occurs. In indirect recognition, NB-LRR proteins are activated by products of the action of Avr proteins on the host. The 'guard hypothesis' has been proposed as a conceptual framework to explain indirect recognition. Pathogen Avr proteins target specific host molecules in order to increase the spread of the pathogen. Host NB-LRR proteins guard these molecules and are activated by changes in their guardees caused by pathogens. NB-LRR proteins either constitutively bind to their guardees and disengage and are activated when pathogen Avr proteins interact with the guardee. Alternatively, NB-LRR proteins are activated by the molecular complex produced when the Avr protein binds with the guardee [6, 7].

As well as the protective mechanisms targeted at pathogens such as bacteria, viruses and fungi just described, plants also possess an array of mechanisms designed to protect them against herbivores such as insects and mammals. These mechanisms are triggered by wounding of the plant by herbivores which causes the production of both direct and indirect defences which are often tailored to the attacking herbivore. Direct defences include the release of antidigestive proteins which reduce the performance of the herbivore by interfering with its digestive enzymes, and the release of antinutritive enzymes which decrease the nutritional value of the plant. Indirect defence mechanisms result in the production of volatile organic compounds (VOCs). These VOCs attract herbivore predators and parasites, and allow top-down control of herbivore populations [8].

Lastly, plants possess systems that are unique among recognition systems in that they produce responses that are the converse of immune responses. Recognition of self (the same plant) produces a response, and nonself (a different plant) does not produce a response [9]. Hermaphroditic plants which produce both pollen and pistel have developed recognition systems to prevent inbreeding, that is, fertilisation of the plant by itself. These self-incompatibility (SI) systems allow plant species to maintain genetic diversity. SI systems depend upon a set of highly polymorphic genes called the S locus which code for both an S-locus receptor protein kinase (SRK) and an S-locus cysteine-rich (SCR) ligand. The SRK receptor is present on the pistel, while the SCR ligand appears on pollen. Binding of SCR to SRK from the same S locus i.e. the same plant, activates the SRK receptor and leads to the arrest of fertilisation, whereas SCR derived from S loci of different plants does not activate the SRK receptor and allows pollination to proceed [10].

## 2.2 Vertebrate and invertebrate immune systems

Around 97% of all animal species are invertebrates and have no adaptive immune system. Yet their immune systems have evolved to help make them the most prolific animals on the planet. Invertebrate immune systems are "not homogeneous, not simple, not well understood" [11]. Studies of invertebrate immune systems have demonstrated that, while only possessing innate immune systems, their immune systems still exhibit phenomena such as specificity, diversity and memory which were previously only associated with the adaptive immune system. For example, in mosquitoes, Down syndrome cell adhesion molecule (Dscam) has been identified as having characteristics similar to human immunoglobulin and is able to produce a diverse set of over 30,000 proteins which enable specific recognition of bacteria. Diversity of Dscam proteins is produced in a similar way to vertebrate immunoglobulin through somatic rearrangement of Dscam gene segments [12, 13]. Invertebrates have also been shown to exhibit specific memory, that is, enhanced protection against the same pathogen upon reinfection. Long-lasting upregulation of regulatory pathways and production of stable proteins such as fibrogen-related proteins (FREPs) in snails have been proposed as mechanisms of specific memory in invertebrates [14, 15].

Evolution has taken different routes to achieve functionally similar systems. In other words, both invertebrate and vertebrate immune systems have evolved different mechanisms which provide antigen-specific memory and protection. Immunoglobulin-based adaptive immune systems have been identified in almost all jawed vertebrates, but not in jawless vertebrates or invertebrates [16]. Lampreys and hagfish, both jawless vertebrates, do not produce immunoglobulin, but instead generate their own diverse set of proteins called variable lymphocyte receptors (VLRs) in response to invading microbes. Lampreys can generate up to 100 trillion unique VLRs. VLRs are made up of proteins called leucine-rich repeat (LRR) modules, and their diversity is generated by a process of somatic rearrangement of LRR modules which surround a single VLR gene [17, 18]. This process of protein rearrangement contrasts with the generation of T cell receptors by somatic recombination of multiple VDJ gene segments in jawed vertebrates. For a review of immune system mechanisms from an evolutionary perspective in invertebrates, protochordates, and jawed and jawless vertebrates see [3].

Thus, while both jawed and jawless vertebrates possess an adaptive immune system, the underlying components and processes of their systems have evolved in different ways. And while invertebrates have no adaptive immune system, they have evolved innate immune systems which provide similar functionality to vertebrate adaptive immune systems. The end result is the same - the production of a diverse set of proteins that provide the host with a mechanism of specific recognition, diversity and memory. The commonalities between Dscam, VLR and immunoglobulin molecules could provide important insights into the essential properties which AISs need to reproduce in their artificial T cell receptors. The differing somatic and germline rearrangement mechanisms which are involved in the generation of Dscam, VLR and immunoglobulin diversity could, for example, provide inspiration for new AIS gene library algorithms.

## 3 Systemic models of the human immune system

Systemic immunological models explore how systemic properties such as immunity and tolerance are generated by the immune system as a whole. The immune system is at this level a system, an assemblage of different interacting entities which comprise a whole. Essentially, systemic models seek to answer questions about what the immune system does and how it does it. Obviously, an un-derstanding of such models is essential for computer scientists seeking to build AISs which exhibit similar systemic properties to the biological immune system. However, the majority of AISs to date have been based on the assumption that the overall purpose of the immune system is to protect the host, and that it does so by mechanisms based around self/nonself discrimination. Adoption of more sophisticated and realistic contemporary models is necessary if AISs are to prove successful at solving hard realworld problems. These models are discussed further in relation to AISs in [19, 20].

Over the course of several decades immunologists have developed a number of systemic models of immunity. For a historical overview and comparison of some of these models see [21, 22]. Many of the more contemporary models are discussed by their protagonists in the internet-based "The Great Debate: The web debate on self-nonself" [23], in which, over a period of five days, leading immunologists debate these models via email and offer some keen insights into their similarities and differences. These models have reflected and guided experimental research. Sakaguchi [24] characterises immunological research in terms of two ancient Greek mottos of Delphi: "Gnothi Seauton" (know thyself) and "Meden Agan" (nothing in excess). He contends that while 'know thyself' has been a favourite slogan of immunologist for many years, the important of 'nothing in excess' has received relatively little attention. The latter truth, manifested in immune homeostasis and self-tolerance, is however, a vital principle of immunity. In this section, we briefly overview these various systemic models and present a categorisation in terms of the way these models view the relationship of the immune system to the body and to itself. In essence, models can be categorised as to whether they see the immune system as a protector or maintainer of the body or of itself. One common feature of contemporary models is the central role they give to the innate immune system as controller of the immune system.

### 3.1 Self/nonself discrimination

It has been long been observed that when pathogens, destructive microorganisms such as viruses or bacteria, enter the body, the immune system removes them and returns the body to a healthy state. Naturally then, the purpose of the immune system is often seen as that of a protector or defender of the body. Since the immune system reacts to pathogens (nonself in immunological terms) but not to the body (self), it also seems logical to conclude that the immune system provides this protection by discriminating self from nonself. Defence by self/nonself discrimination has formed the basis of the majority of immunological models since the middle of the last century, and this view of the immune system is still widely accepted by immunologists today [25].

Earlier models of immunity were based around the idea that host constituents (self) are ignored by the immune system, while other elements (nonself), such as pathogens, foreign substances or altered self, are reacted to. In these models, tolerance is largely viewed as immune system silence or nonreactivity to self. Models such as Burnet's Clonal Selection Theory [26] and the Associative Recognition Model of Bretscher and Cohn [27] rest on a historical mechanism in which immature receptor-bearing cells of the adaptive immune system are exposed to a wide range of non-pathogenic material early in the development of the organism. If this non-pathogenic material is recognised above a certain level, this leads to the destruction of the cell and its receptors. This results in a set of mature cells whose receptors only recognise antigen which are not historically part of the organism. This recognition leads to the initiation of an immune response and the destruction of the pathogen to which the antigen belongs.

## 3.2 Infection and danger

Other models, based on divisions of antigen other than self/nonself, have also been developed. For these models, the immune system does not partition the antigenic universe into two groups of self and nonself molecules. Self/nonself discrimination has been criticised for being applied to the mechanisms which produce overall immune system behaviour. It has been observed [28] that referring to self/nonself discrimination of antigen by T cells is a category error. While the immune system as a whole appears to recognise self from nonself (a systemic property) this does not imply that individual T cells recognise self antigen. This is making the mistake of attributing the property of a system to its elements. While preserving the protective purpose of the immune system, models such as the Infectious Nonself Model and Danger Model have moved away from self/nonself discrimination as the driving force behind immunity.

The Infectious Nonself Model of Janeway [29] like earlier models, views the purpose of the immune system as protecting the body. However, the Infectious Nonself Model proposes that instead of categorising antigen into self and nonself, the immune system categorises antigen into the classes of infectious nonself and noninfectious self. Moreover, instead of the adaptive immune system based historical process of negative selection, detection of pathogens by innate immune system cells is seen as the principal controller of the immune system. Janeway proposes that innate immune system antigen presenting cells (APCs), especially dendritic cells, are the principal controllers of the immune system. In a similar way to plant cells as discussed in Section 2.1, APCs express a groups of receptors called pattern recognition receptors (PRRs) which respond to pathogen-associated molecular patterns (PAMPs). Janeway defines PAMPs as "conserved molecular patterns that are essential products of microbial physiology ... unique to microbes ... [and] not produced by the host ... [which] are recognized by receptors of the innate immune system ... [and which] induce the expression of costimulatory molecules on the cell [APC] surface, which is necessary for the activation of naive T cells" [30]. Activation of PRRs by PAMPs initiates and modulates an immune response, and the activation of different subsets of PRRs tailors the immune response to different classes of pathogens.

The Danger Model of Matzinger [31, 32] is similar to the Infectious Nonself Model, viewing the immune system as a protector, and the innate immune system as having a central role in the generation of protection. It also agrees that APCs have PRR receptors which when bound to certain molecules, activate the APC, allowing it to express antigen in a stimulatory fashion. However, instead of being specific for material associated with pathogens, these receptors are specific for molecules, termed danger signals, produced when the tissue of the organism is damaged or stressed. Matzinger defines danger signals as "a set of molecules elaborated or released by stressed or damaged cells, for which resting APCs have receptors, and to which resting APCs respond by becoming activated and upregulating costimulatory capacity " [33]. Danger signals are released by cells when they undergo necrosis, unprogrammed death, but not when they undergo apoptosis, cell death which occurs as part of the normal functioning of the organism.

Although both models appear similar, their explanations of the origin of the material which activates APCs, and hence is responsible for the activation of an immune response, lead to important differences. By proposing host damage as the main regulator of the immune system the Danger Model expands the definition of the innate immune system to include the tissue cells of the host itself. In fact, these tissue cells are the cells that control innate immunity, and the class of immune response is not determined by the pathogen but rather by these tissue cells themselves. A key similarity between these models is the shift in control of the immune system from the T and B cells of the adaptive immune system to the cells of the innate immune system.

### 3.3 Maintenance and homeostasis

Notions of dangerous and harmless are, however, themselves problematic. Philosophers such as Canguilhem [34] and Haraway [35] have observed how the concepts of pathological and normal are just as metaphorical and observer-dependant as those of self and nonself. Is the pathological an overexpression of the normal (a hyperreaction) or is it a radically different state from the normal? What exactly does the normal or average state mean? How have wider social and scientific notions of self and nonself influenced the way the immune system is understood? Other models have emerged which challenge the view of the immune system as a purely defensive and discriminatory system, and widen its functions to include host-maintenance and self-assertion or homeostasis.

Models such as these generally reject the notion that recognition equals pathogenicity. Instead, there is constant recognition and reaction by the immune system, which leads to a tolerogenic or immunogenic response. In these models, in place of a defender, the immune system is viewed as a maintenance or homeostatic system, maintaining the body or itself respectively. Resistance to change, for example produced by a pathogen, results in behaviour that appears to protect the body and recognise the pathogen. But it is the maintenance of the body in a particular state that is really the driving force behind this behaviour. Some models go further and assert not only that the purpose of the immune system is maintenance, but that it is self-maintenance or self-assertion, and not body maintenance. These models view the immune system as a homeostatic system, an open system which regulates its internal environment and maintains a state of dynamic equilibrium in the face of changes in its environment.

Cohen's cognitive paradigm [36, 37] describes the immune system as a cognitive systems in which a dialogue is constantly taking place between immune cells and the body. Interactions between, for example, APCs and T cells can be described in terms of APCs communicating sentences describing the nature of an antigen to T cells. The subject of the sentence is the antigen. The predicate is a complex set of costimulatory molecules and cytokines produced by the surrounding tissue, or by APCs in response to signalling through germline innate receptors. The immune meaning of an antigen is defined as how the T cells responds to this sentence, with the context of the antigenic subject provided by the predicate. Through continuous dialogue between immune cells and the host, the immune system generates an internal image of self, which Cohen terms the immunological homunculus. Andrews and Timmis discuss the cognitive paradigm in relation to AISs further in [38].

The Morphostasis Model of Cunliffe [39, 40] is based on the idea that the function of the immune system is tissue homeostasis. All cells in the body are able to sense when their normal function is disrupted, for example through co-option by a virus. When this occurs, cells signal this abnormality to neighbouring cells and sometimes apoptose. However, many pathogens have developed the ability to prevent their target cells from apoptosing. Phagocytic innate immune system cells such as macrophages and neutrophils play a key role in the Morphostasis Model. Phagocytes are able to sense changes in the normal functioning of cells and remove these cells. In this model the role of the adaptive immune system is to accelerate the identification and clearance of non-healthy cells by phagocytes. The Integrity Model of Dembic [41, 42] is similar to the Morphostasis Model in that it characterises the immune system as maintaining the body through surveillance of the state of tissue. In the Integrity Model innate immune system dendritic cells scan tissue and detect changes in signal levels produced by tissue cells. This induces dendritic cells to initiate an adaptive immune system response.

## 4 Conclusion

In this paper we have argued for the need for AISs which are based on much more biologically-realistic models. We argued that, instead of building AISs based on the extremely complex human adaptive immune system, AISs should draw inspiration from relatively simpler organisms which possess only innate immune systems. In the first half of this paper we outlined current biological understanding of plant and invertebrate immune systems. The innate immune systems of these organisms are capable of self/nonself discrimination, and also exhibit properties such as specificity, diversity and memory which until recently have only been associated with adaptive immune systems. Low-level biological models of the mechanisms which give rise to these properties could provide important sources of inspiration for future AIS algorithms. If AISs are however to employ adaptive immune system mechanisms, then we argued that they also need to incorporate innate immune system mechanisms, which control the adaptive immune system in biological organisms. In the second half of this paper we outlined a number of systemic models of the human immune system which deal with how the innate and adaptive immune systems are integrated. These models provide AIS design-ers with a concrete framework for incorporating innate and adaptive immune mechanisms into their artificial systems.

As well as producing more effective AISs, building AISs based on more biologically-realistic models could have important consequences for biological research. For a number of years we have had the opportunity to work closely with immunologists. During this time we have been keen to develop interdisciplinary relationships which have benefited these immunologist as much as they have benefited us. Realistically however, this has proved very difficult, and we feel that we, and the field of AISs in general, have had very little impact on immunological research and thinking. Part of the reason for this is that the naive models employed by AISs have borne little resemblance to the models of immune system mechanisms employed by immunologists. Perhaps a more fundamental reason for this state of affairs is that the focus of immunological and AIS research often differ. Immunology has been largely focussed on elucidating the cellular and molecular basis of the immune system using a reductionist methodology. The field of AISs on the other hand is often concerned with building complete systems and adopts a more holistic methodology. An exception to this within Immunology is Systems Immunology, which studies how entities and mechanisms interact at different system levels to determine immune system behaviour, and whose domain includes systemic models of the immune system. Here we believe that AISs, by building artificial systems based on more biologically-realistic systemic models of the immune system, could have a significant impact. Such AISs, when applied to complex realworld problems, could provide important experimental systems which could be more easily manipulated and from which data could be more easily gathered than biological systems. These AISs could then be used to validate systemic immune system models. In this way, AIS research could have a real impact on Immunology.

AIS research to date has largely been concerned with engineering, that is, building useful machines or systems which solve practical problems. Whether or not the immune-inspired principles used reflect any fundamental properties of biological immune systems is of little consequence. If they are useful in achieving the practical ends of the engineer then they have served their purpose well. This engineering approach, in our opinion, while being productive in developing solutions to practical problems, has further limited the interdisciplinary impact of AIS research. What is needed to address this limitation is an expansion of the scope of AIS research to address fundamental questions in Immunology and the organisation of complex systems. This can only be done if, on the one hand, more biologically-realistic models are adopted and studied specifically to understand the dynamics of these models, and whether they capture the dynamics of the biological systems they seek to describe. On the other hand, insights from computer science into complex systems and techniques for modelling such systems could help immunologists to develop better biological models. Perhaps what we will see in the future is AISs grow into a field which takes its biology as seriously as its engineering. In this case, a more appropriate definition of the field would be Artificial Immunology - the construction and study of immune-systems-as-they-could-be in an effort to understand immune-systems-as-they-are and to enhance the construction of immune systems for artificial organisms.

## Acknowledgements

This research is supported by EPSRC Grant GR/S47809/01.